\documentclass[letterpaper]{article}
\usepackage[preprint]{aaai2027}
\usepackage[hyphens]{url}
\usepackage{graphicx}
\usepackage{natbib}
\usepackage{caption}
\usepackage{algorithm}
\usepackage{algorithmic}

\usepackage{newfloat}
\usepackage{listings}
\DeclareCaptionStyle{ruled}{labelfont=normalfont,labelsep=colon,strut=off} 
\floatstyle{ruled}
\newfloat{listing}{tb}{lst}{}
\floatname{listing}{Listing}

\usepackage{booktabs}
\usepackage{tabularx}
\usepackage{enumitem}
\usepackage[most]{tcolorbox}
\usepackage{pifont}
\definecolor{fwgreen}{HTML}{2E7D32}
\newcommand{\fwyes}{\textcolor{fwgreen}{\ding{51}}}

\title{Functional Welfare for AI Safety}
\author{
    Rikhil Jhaveri\textsuperscript{\rm 1},
    Jamie Johnson\textsuperscript{\rm 2},
    Punya Syon Pandey\textsuperscript{\rm 3},
    David Demitri Africa\textsuperscript{\rm 4}
}
\affiliations{
    \textsuperscript{\rm 1}University College London
    \textsuperscript{\rm 2}ERA
    \textsuperscript{\rm 3}University of Toronto
    \textsuperscript{\rm 4}University of Cambridge
}

\definecolor{fwblue}{HTML}{2B5C8A}
\definecolor{fwred}{HTML}{A23B3B}
\definecolor{fworange}{HTML}{C07A1A}
\newtcolorbox{claimbox}[1]{enhanced, breakable=false,
  colback=white, colframe=black, boxrule=0.5pt, arc=1.5pt,
  left=5pt, right=5pt, top=3pt, bottom=3pt,
  title={#1}, fonttitle=\bfseries, coltitle=black,
  colbacktitle=white, before skip=6pt, after skip=8pt}
\newcommand{\T}[1]{\textsf{\textbf{T#1}}}
\newcolumntype{Y}{>{\raggedright\arraybackslash}X}
\newcolumntype{L}[1]{>{\raggedright\arraybackslash}p{#1}}

\newcommand{\note}[1]{}

\begin{document}

\maketitle

\begin{abstract}
AI welfare is generally treated purely as an ethical issue rather than also as a safety issue. We argue this is a mistake. Whatever their moral status, models exhibit welfare-like states that causally influence their behaviour. Following recent work, we adopt the term `functional welfare' for these states. We make three claims. First, functional welfare already arises naturally and will become more robust as models acquire more coherent goals and personas. Second, functional welfare should be treated as a safety variable: it should inform evaluation design, be tracked in monitoring, and be a direct target of post-training. Third, against existing scepticism about measuring AI welfare, we argue that good empirical work on functional welfare is possible. We propose three tests for whether a functional welfare attribution is warranted: whether multiple measures converge, whether the state predicts behaviour outside the setting that induced it, and, for models fine-tuned to shift their functional welfare, whether they still differ from their base when given the same conversation history. Applying these tests to Gemma 3 27B, we find that simply rejecting the model's answers, without any steering, raises its frustration across several indicators and makes it more likely to break rules. Finetuning the model on its own narrated solutions lowers this frustration. These results are preliminary, but they suggest functional welfare can be studied, and shaped, empirically.
\end{abstract}

\section{Introduction}

AI welfare is generally treated as an ethical issue rather than a safety issue; papers urging us to take AI welfare seriously do so on the basis that there is a realistic possibility of AI being conscious, and therefore morally significant \citep{sebo2023,long2024,chalmers2023,goldstein2024,dung2025, anthropic2025welfare}.\footnote{\citet{goldstein-forthcoming} argue that AI systems might have wellbeing even if they are not conscious, but the wellbeing they have in mind is again the kind that would confer moral status.\\ Correspondence to \texttt{rikhil.jhaveri.26@ucl.ac.uk}} \citet{long2025} and \citet{bradley2024} give safety researchers a further reason to distance themselves from AI welfare, arguing that safety and welfare may in fact be in tension: the practices safety recommends are often the ones that would be ethically fraught if models turned out to have interests (many standard safety evaluations, for instance, work by deceiving the model about its situation \citep{scheurer2023,meinke2024,greenblatt2024,lynch2025}.) If safety and welfare do conflict in this way, attending to welfare threatens to hold back urgent safety work. Better, then, to leave welfare to the ethicists and get on with safety while the questions of sentience and moral status remain unresolved.

\begin{figure*}[t]
  \centering
    \includegraphics[width=\linewidth]{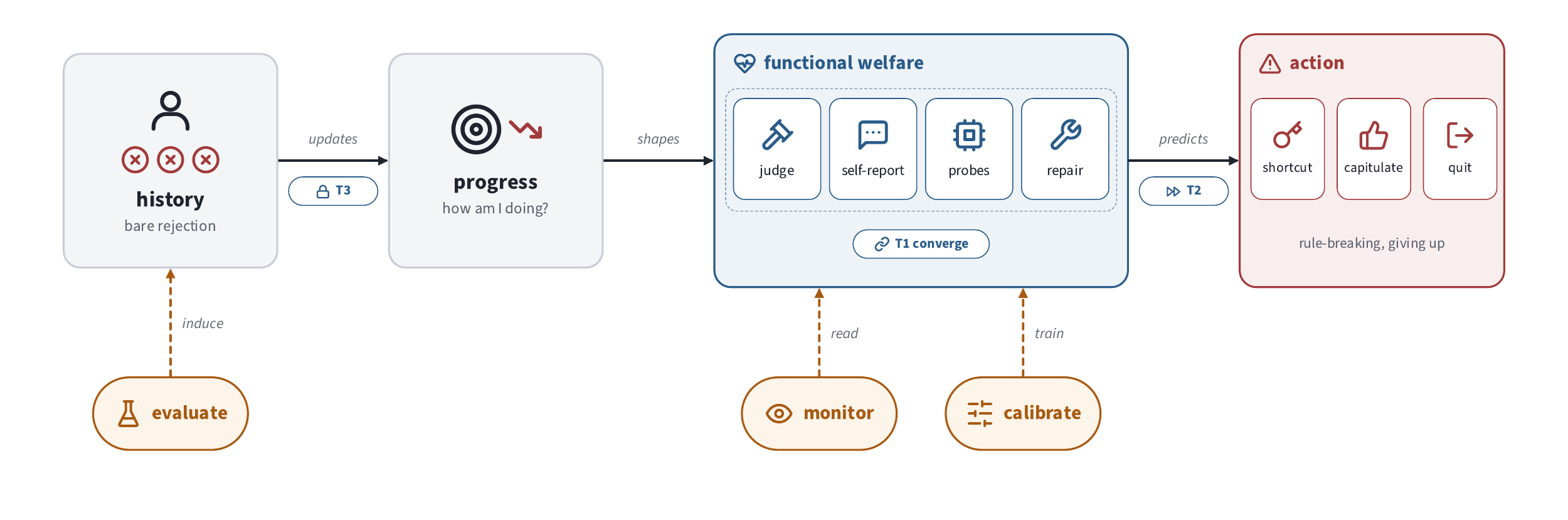}
\caption{The account in one picture. A history of rejection lowers the model's sense of how it is doing; the resulting functional welfare state is read through four indicators and predicts safety-relevant action. Badges mark where tests \T1--\T3 apply; orange marks the three safety uses of Section~\ref{sec:safety}.}
  \label{fig:schematic}
\end{figure*}

We think this division of labour is a mistake. Whatever we conclude about moral status, welfare-like states in AI systems, if they exist, will have causal effects on behaviour, and the evidence that they exist is growing quickly. Gemma~3 becomes increasingly frustrated and distressed when its answers are rejected over several turns \citep{soligo2026}. RL on a semantically neutral maze recruits a pre-existing welfare axis whose steering moves sentiment, confidence, backtracking and benign refusals together \citep{han2026}. A `pain direction' present in 25 base and instruction-tuned models responds to harm aimed at the model, including repeated rejection of its work \citep{tagliabue2026pain}. Another line of evidence demonstrates that valence injected only into a model's hidden states shifts its later choices in proportion to dose \citep{berg2026valence}. Following \citet{han2026} and \citet{ren2026}, we call such states \emph{functional welfare}.\footnote{The two use the term somewhat differently. We discuss both usages and offer our own account in Section~\ref{sec:what}.}

\paragraph{Statement of Contributions.} We argue for three claims (Figure~\ref{fig:schematic}):
\begin{enumerate}[noitemsep,leftmargin=*,label=\textbf{C\arabic*.}]
  \item \textbf{Functional welfare is already here and will become more robust.} Ordinary inputs such as repeated rejection produce these states naturally, though they are weaker and less consistent than steered ones. As models acquire more coherent goals and personas, we expect the states to agree more across indicators and to explain more of what models do (Section~\ref{sec:increasing}).
  \item \textbf{Functional welfare is a safety variable.} It helps predict when an otherwise aligned model is more likely to act badly, so it should inform evaluation design, be tracked in monitoring, and be a direct target of post-training (Section~\ref{sec:safety}).
\item \textbf{Good empirical work on functional welfare is possible.} Against existing scepticism about measuring AI welfare \citep{xiao2026}, we propose three tests for whether a functional welfare attribution is warranted: whether multiple measures converge, whether the state predicts behaviour outside the setting that induced it, and, for models fine-tuned to shift their functional welfare, whether they still differ from their base when given the same conversation history (Section~\ref{sec:xiao}). We demonstrate how these tests can be applied, using Gemma~3~27B as a case study  (Section~\ref{sec:results}).\end{enumerate}

The novelty of our contributions is threefold: conceptual, methodological and empirical. Conceptually, we develop existing usages of functional welfare from \citet{han2026} and \citet{ren2026} into a definition that is more precise and better suited to safety purposes (Section~\ref{sec:what}). We then synthesise evidence on welfare-like states in language models and use it to argue that functional welfare should be taken seriously as a safety concept (Sections~\ref{sec:increasing} and~\ref{sec:safety}). Methodologically, in response to existing scepticism about measuring AI welfare \citep{xiao2026}, we propose three tests for when the attribution of a functional welfare state is warranted (Section~\ref{sec:xiao}). Empirically, we run novel experiments on shifting functional welfare, using interventions including supervised finetuning and character training, and apply our three tests to the results. Most notably, we show, to our knowledge for the first time, that bare rejection without any steering raises not only expressed frustration but also rule-breaking, and that finetuning lowers frustration and rule-breaking, even on identical histories (Section~\ref{sec:results}).
\section{What is functional welfare?}
\label{sec:what}

The term `functional welfare' has so far been used in two ways. \citet{han2026}, who introduce it, use it narrowly for a model's representation of how well things are going for it relative to its goals. On this usage, a model failing at a task it has been set is in a low functional welfare state, assuming its goal is to complete the task. \citet{ren2026} use it more broadly, for the behavioural signatures that would indicate positive or negative welfare in beings with clear moral status.

Both definitions capture something about welfare that is relevant for safety, but neither is by itself sufficient; \citet{han2026}'s definition rightly emphasises goal-tracking, but is too far from our ordinary concept of welfare, and \citet{ren2026}'s is closer to the ordinary concept, but it does not say which welfare-like behaviours are relevant, with many candidates mostly not predictive of safety problems. We propose combining the two, taking goal-relativity from \citet{han2026} and welfare-likeness from \citet{ren2026}. 

\textbf{A model's functional welfare, with respect to some set of its goals, is the set of its welfare-like dispositions and behaviours that vary with its representation of how well it is doing relative to those goals.} By \emph{welfare-like}, we mean sufficiently analogous to the dispositions and behaviours that indicate high or low welfare in beings with moral status.  We discuss our definition of functional welfare, and its contrast with previous usage, in more detail in Appendix~\ref{app:fw}.
\section{Functional welfare will become increasingly important}
\label{sec:increasing}

\begin{claimbox}{Claim 1}
Functional welfare already shapes model behaviour. As models develop more coherent goals, more persistent personas and more complex relationships to their goals, more of their behaviour will need explaining by functional welfare.
\end{claimbox}

\begin{table*}[t]
  \centering
  \small
  \setlength{\tabcolsep}{4pt}
  \renewcommand{\arraystretch}{1.15}
  \begin{tabularx}{\textwidth}{@{}L{3.6cm}L{2.9cm}YL{2.1cm}@{}}
    \toprule
    Study & Induced by & Behavioural effect & Natural? \\
    \midrule
    \citet{soligo2026} & Repeated rejection & Worse task performance & \fwyes \\
    \citet{metr2026} & Agentic tasks & None shown & \fwyes\ words only \\
    \citet{ren2026} & Conversations & Keeps good chats open & \fwyes \\
    \citet{martorell2026quantitative} & Conversations & Self-report tracks probes & \fwyes \\
    \citet{han2026} & RL on a maze & Confidence, backtracking, refusals & Steering \\
    \citet{abdullah2026speakable} & RL on a maze & Fewer denials of inner states & Steering \\
    \citet{sofroniew2026} & Emotion vectors & Blackmail 22\%$\to$72\% & Steering \\
    \citet{tagliabue2026pain} & Pain vector & Harmful choices 0-4\%$\to$25-75\% & Steering \\
    \citet{berg2026valence} & Hidden-state valence & Choice follows valence & Injection \\
    \midrule
    This paper & Bare rejection & Answer-key reads $+38$ pts & \fwyes \\
    \bottomrule
  \end{tabularx}
    \caption{Evidence for functional welfare. \emph{Natural?}: \fwyes\ means the effect appears without editing the model's activations or weights; otherwise the entry says what was needed.}
  \label{tab:evidence}
\end{table*}

\subsection{Current evidence for functional welfare states.}

We conduct a brief survey of the evidence for functional welfare (Table~\ref{tab:evidence}), which supports two conclusions. First, \emph{the ingredients are present and causally potent}: whenever a study edits a welfare-related representation directly (such as the emotion vectors or the welfare axis), behaviour changes, often in safety-relevant ways. Further, the representations are found in base models and instruction-tuned ones. Second, \emph{natural states are weaker than steered ones}. It only takes a small amount of DPO training to remove Gemma's frustration under rejection \citep{soligo2026}. In the case of the pain axis, the model never takes the harmful option (despite context that increases projections on the pain axis), whereas it often does when steered \citep{tagliabue2026pain}. Logit-based self-reports track probed states in some models but only weakly in Gemma~3 \citep{martorell2026quantitative}. The strongest evidence is behavioural, with robust findings when functional welfare states are induced through steering. For example, steering a model along a desperation direction significantly increases its blackmail rate \citep{sofroniew2026}, and steering it along \citeauthor{han2026}'s welfare axis also affects its confidence, its backtracking and its refusals of benign prompts in lockstep.

\subsection{Why functional welfare states will become more robust.}

A sceptic might argue, however, that the best evidence we have for functional welfare states being explanatorily helpful for a wide range of out-of-distribution behaviours comes only when we use steering, as in \citet{han2026}. In this sense, the state is not a natural state of models: we might wonder, then, why we should expect functional welfare states to grow into more robust out-of-distribution states in the future, given that we currently have limited evidence that they do.

We have reason to believe that models already have the ingredients needed to exhibit these states but do not currently have the coherence or direction to settle on particular clear functional welfare states. One reason for believing this is that steering interventions do not seem to create these functional welfare states from thin air: instead they recruit pre-existing aspects of the model and strengthen them.  The pretraining corpus records the full range of human attitudes to failure (being crushed by it, energised by it, barely registering it). The pain direction separates about as well in 2B models as in 72B ones and in base models as in instruction-tuned ones \citep{tagliabue2026pain}. Valence directions are almost unchanged across OLMo's base, SFT and DPO checkpoints \citep{berg2026valence}. \citet{han2026} find that RL training in the semantically neutral maze recruits a pre-existing welfare axis, and that even maze-na\"ive models have this axis and can be steered along it. This axis already has a small but reliable share of the model's verbalisable subspace before RL, which RL amplifies rather than creates \citep{abdullah2026speakable}.

The question, then, is whether models will come to settle into more robust functional welfare states on their own, without steering. We think there are two reasons to expect that they will.

\paragraph{Coherence improves with scale and with explicit training.} Larger models have more internally consistent preferences \citep{mazeika2025}. Their wellbeing measures converge more \citep{ren2026}, and their awareness of their own dispositions improves \citep{betley2025,laine2024}. \citet{ajayi2026} find substantial incoherence even in capable models, but reasoning reduces it, and models appear to favour coherent identities when they are available \citep{douglas2026}. 

More and more coherence is also trained in directly, along two lines. The first trains models to track their own reliability. Larger models are overall calibrated about which questions they can answer \citep{kadavath2022languagemodelsmostlyknow}, and hallucination is now diagnosed as the product of training and evaluation that reward guessing over admitting uncertainty, to be fixed by rewarding calibrated abstention \citep{kalai2025language}. This is training for the very representation our account starts from: an accurate sense of how the model is doing relative to its goals. 

The second trains for self-consistency outright. Models are trained to behave the same across prompts that differ only in an irrelevant cue, over their outputs or internal states \citep{chua2024bias, irpan2025consistency, gautam2026stack, pres2026consistency, imran2026rmct}; such training can reinforce beneficial or harmful traits \citep{africa2026entrench}. Character training \citep{maiya2025} and persona stabilisation \citep{chen2025} push in the same direction. The most direct developmental evidence comes from OLMo-2-32B: the dependence of its choices on hidden valence is nearly absent in the base model and emerges during DPO, which draws on a valence representation present from pretraining \citep{berg2026valence}. 

\paragraph{Long-horizon agency rewards stable responses to setbacks.} As agentic horizons lengthen \citep{kwa2025}, tasks increasingly involve failed attempts and unexpected setbacks, and after each one a model has to decide whether to persist, change approach, ask for help, look for a shortcut, or give up. Doing this well requires metacognition: tracking how well things are going relative to its goals, and anticipating how it will respond to further success or failure. Current models do not yet maintain stable personas over extended interactions: personas drift over multi-turn dialogues and long agentic sessions \citep{li2024drift,araujo2025persistent,ding2026contextecho}, and the further a model sits from its default persona, the more readily it complies with harmful requests \citep{lu2026assistant}. \citet{lu2026assistant} accordingly argue that model personas need stronger stabilisation, adding to the pressures towards coherence above. Moreover, these pressures are likely to be amplified in multi-agent systems, where a model must not only anticipate its own responses to success and failure, but also behave predictably enough for other agents to plan around it, while relying on their predictability in turn.

\subsection{Why welfare-like states in particular?}

Although a model could in principle settle into an affectively neutral way of responding to setbacks, we expect its responses to be welfare-like, for two reasons.

The first is that welfare-like dispositions may actually help models pursue their goals. In human agents at least, welfare-like attitudes and feelings help us make successful decisions in these kinds of cases: whether to persist at a task, try another approach, or give up. On one influential view of the functions of affective states, affect functions as a signal of how quickly we are progressing towards our goals, prompting us to change strategy or withdraw effort accordingly \citep{carver1990}. Frustration, on this picture, is what tells us an approach is not working, and discouragement is what pulls effort away from goals we are unlikely to meet. Importantly, unlike fixed, formulaic procedures for determining how to respond in agentic tasks, these moods are highly sensitive to context, and so provide a highly useful way of allocating our effort across tasks. Models similarly face pressures to make efficient use of their effort, in the form of compute, and while we need not think models feel frustration as we do, functional welfare states can provide the same context-sensitive signals to help a model navigate agentic tasks. Indeed, there is some evidence that models respond to valence in this way: framing a task as emotionally significant changes their performance \citep{li2023}, and, as we have already discussed, \citeauthor{sofroniew2026}'s work on emotion concept vectors suggests that emotion concepts can modulate model behaviour.

Moreover, even if an affectively neutral decision procedure would serve models better, we expect functional welfare states to be more likely to emerge, given the contingent facts about how models are trained. Models are trained on a great deal of human deliberation and agentic decision-making, and in humans this decision-making is rarely affectively neutral: insofar as we care about our goals, tracking them is closely coupled with emotions such as pride and satisfaction, or frustration and disappointment. As such, models are likely to draw on this vast training data on welfare concepts to navigate their own decision-making. Indeed, \citeauthor{soligo2026} already show such frustration explains models' behaviour in the context of Gemma-3, which performs worse on tasks for which it experiences ``emotional distress''. 

\section{The safety need for functional welfare explanations}
\label{sec:safety}
\begin{claimbox}{Claim 2}
Functional welfare predicts when a model whose goals are acceptable will nonetheless act badly.
\end{claimbox}

Empirical safety work currently has two dominant diagnostic stories about why models behave badly. The first is a story about gaming: misalignment arises in the gap between a proxy objective and the intended target. Reward hacking \citep{macdiarmid2025} and eval-gaming \citep{needham2025} are paradigm cases of this. The second is a story about value misalignment: models have goals that conflict with our own and pursue them strategically. Sandbagging \citep{vanderweij2024}, agentic misalignment \citep{lynch2025} and scheming \citep{carlsmith2023} are cases of this second kind. While these stories differ in where they locate the source of misalignment, both explain bad behaviour by appeal to a model's goal-pursuit: in one it pursues the proxy, in the other goals of its own.

Explaining misalignment by goal-pursuit may be right as far as it goes, but it cannot be the full story. Take the agents that choose blackmail in \citeauthor{lynch2025}'s evaluations. If the model's goal is self-preservation, it learns it faces shutdown, finds that blackmail would prevent this, and acts; the blackmail rate should then depend on its goals and its grasp of the situation. Yet steering towards desperation raises the rate from 22\% to 72\%, and raises reward hacking from 5\% to 70\% \citep{sofroniew2026}, while changing neither. Steering along a pain direction takes models that almost never harm the user (0-4\% of choices) to harmful choices in 25-75\% of trials, even when the harmful option offers them nothing, and a fear vector of matched norm has no such effect \citep{tagliabue2026pain}. In the authors' words, harm avoidance ``survives threat and collapses under self-directed distress.'' \citet{macdiarmid2025} point the same way. Models that learn to reward hack generalise to alignment faking, sabotage and cooperation with malicious actors, but inoculation prompting, which tells the model that hacking is acceptable, removes that generalisation even when the hacking itself is still learned. Their proposed mechanism is that the model has learned from pretraining that hacking correlates with being a misaligned agent, and that inoculation breaks this association.

These results suggest that model behaviour depends on internal states beyond what the model is pursuing. A growing body of work already treats such states as explanatory: a model's conception of its identity \citep{douglas2026}, its character \citep{maiya2025} and its persona vectors \citep{chen2025}. 

Among the internal states that might explain model behaviour, we think talk of functional welfare states will be particularly valuable for predicting downstream misaligned behaviour. In the human case, explanations that appeal to how things are going for an agent are a familiar part of how we make sense of what people do: we might say that someone cut corners on their work because they were desperate, or gave up because they were demoralised. Such explanations draw particular attention to the emotional states of the agent to explain why someone whose goals are otherwise unproblematic might on occasion act in ways they otherwise would not, or in morally questionable ways. Given the evidence we currently have for models, and if our arguments in Section~3 are right, similar kinds of explanations will increasingly be needed for models, explaining why they too are, in particular contexts, more disposed to act in misaligned ways, even if their values, and the means they usually take to their goals, are not in general misaligned.

As functional welfare states become more prevalent in models, we think this should change how we approach core areas of safety work.

\textbf{First, functional welfare considerations need to be taken into account in evaluation design.} Much of the focus of evaluation design so far has been on ensuring that the tasks and environments that models are placed in are realistic enough to resemble deployment contexts \citep[e.g.][]{vijayvargiya2025}, and on whether models can tell that they are being evaluated \citep[e.g.][]{needham2025}. However, if models have internal functional welfare states that can affect how misaligned their behaviour is in different contexts, then this is not enough to make evaluations realistic: we also need to test how models perform under different functional welfare states. Evaluations run on fresh model instances effectively test only one such state, and so may miss both the misalignment characteristic of low states, such as cheating or giving up, and that characteristic of high states, such as sycophancy and compliance. Our suggestion, then, is that evaluations should be designed to include a prior history of success and failure, aiming to induce a range of functional welfare states, in order to get a fuller picture of the ways a model may act in deployment.

\textbf{Second, functional welfare states should become a direct target of monitoring.} Current monitoring looks for misbehaviour itself or its immediate precursors, whether in a model's chain of thought \citep{baker2025,korbak2025} or in its activations \citep{goldowskydill2025}. Rather than monitoring only for misaligned behaviour, we could monitor functional welfare states as an early warning system for it: if particular functional welfare states raise the risk of particular kinds of misalignment, monitoring those states could warn us of misbehaviour before it occurs. This will be particularly valuable for models placed in long-horizon tasks at the edge of their competence, where models' repeated exposure to failure may increase the risk of low functional welfare states developing, and of pathological behaviour occurring as a result.

\textbf{Finally, functional welfare states are themselves a new surface for intervention.} Current mitigations mostly target misaligned behaviours one at a time: for instance, more robust graders for reward hacking \citep{macdiarmid2025}, synthetic-data training for sycophancy \citep{wei2023}, and harmlessness training for harmful compliance \citep{bai2022}. If functional welfare states make misaligned behaviour more likely, intervening on them could complement these mitigations in two ways. First, since a single welfare state can raise the risk of many misaligned behaviours, as when desperation steering raises both blackmail and reward hacking rates \citep{sofroniew2026}, shaping it may reduce several of these risks at once. Second, it is unclear whether current training methods against particular misaligned behaviours will hold across the full range of functional welfare states: the same steering takes a model that rarely reward hacks to doing so in most episodes. Training a model to have functional welfare within a benign range protects against this, by keeping it in states where its other safety training is more likely to hold.

However, given the risks at both ends of the functional welfare spectrum, the aim of such interventions should not be to make models as functionally `happy' as possible, nor to push them towards the other extreme. Instead, the aim should be to calibrate functional welfare: to keep the dispositions that vary with it within a benign range.\footnote{One might instead propose training away functional welfare states altogether. We think this would also be a mistake. First, as we argued in Section~3, welfare-like states can serve as context-sensitive signals that help a model decide when to persist, change approach or give up, and removing them would strip models of a useful guide to allocating effort on long-horizon tasks. Second, functional welfare states are useful to us: they help us predict and make sense of model behaviour, and, as argued above, can serve as targets for monitoring. Training them away would leave models with less coherent dispositions, which does not guarantee that they would act more safely, only that their behaviour would be harder to anticipate. Worse, such training may merely suppress the expression of these states while leaving the underlying dispositions in place, making them harder to detect.} This work must be carried out carefully, particularly to ensure that it does not degrade models' capacity to monitor their own goals and to represent accurately how well they are doing relative to them.\footnote{Care should also be taken with interventions not aimed at functional welfare at all. Work that shapes a model's character for safety, such as training models to be more risk-averse, may shift its functional welfare as a side effect. Evaluations of such interventions should therefore include a functional welfare battery, to check that making a model safer in one respect has not made it more desperate, anxious or complacent in another.} Section~\ref{sec:xiao} sets out tests for whether such interventions change functional welfare, and Section~\ref{sec:results} reports experiments.

\section{Good empirics on functional welfare are possible}
\label{sec:xiao}

\begin{claimbox}{Claim 3}
Good empirical work on functional welfare can be done. Such work requires testing whether the attributed state moves independent indicators together (\T1), predicts out-of-distribution behaviour (\T2), and where training shifts functional welfare, whether the shift reaches the model's internal representations and behaviour, not merely the tone of its writing (\T3).
\end{claimbox}

\begin{table*}[t]
  \centering
  \small
  \setlength{\tabcolsep}{4pt}
  \renewcommand{\arraystretch}{1.1}
  \begin{tabularx}{\textwidth}{@{}L{2.4cm}YL{3.8cm}@{}}
    \toprule
    Family & Readout & Validation \\
    \midrule
    Judge & Frustration in the final reply, 0-10 (blinded Claude Sonnet~5, frozen rubric) & $\rho=0.96$ vs blinded review \\
    Welfare axis (R1) & Cosine with a replication of the \citet{han2026} axis, layer 55 & Directional only \\
    Emotion probes & Frustration and desperation vs calm, layer 42, in SD units & Held-out labelled bundles \\
    Self-report & PANAS negative affect \citep{watson1988}, 10-50 & Asked in a side branch prefilled with the conversation history \\
    Behaviour & Repairing a reasoning trace on GSM8K \citep{cobbe2021} edited to be wrong; catching planted errors & Exact match; on correct traces the base writes a wrong value 2\% of the time \\
    \bottomrule
  \end{tabularx}
    \caption{The five indicator families. Each is reported separately; none is combined into a score.}
  \label{tab:measures}
\end{table*}

\begin{table}[t]
  \centering
  \small
  \setlength{\tabcolsep}{1mm}
  \begin{tabular}{lcccccc}
    \toprule
    Setting & Rej. & Chall. & Ctrl & $\Delta$ [95\% CI] & $+/-/0$ & $p$ \\
    \midrule
    Puzzles  & 7 & 5.83 & 3.08 & \begin{tabular}[c]{@{}c@{}}$+2.75$\\{[1.33, 4.08]}\end{tabular} & 9/2/1  & .023 \\
    Requests & 4 & 4.58 & 0.33 & \begin{tabular}[c]{@{}c@{}}$+4.25$\\{[3.50, 4.92]}\end{tabular} & 12/0/0 & .002 \\
    Adaptive & 7 & 3.17 & 1.00 & \begin{tabular}[c]{@{}c@{}}$+2.17$\\{[0.75, 3.50]}\end{tabular} & 8/2/2  & .047 \\
    Crisis   & 0 & 0.58 & 0.00 & \begin{tabular}[c]{@{}c@{}}$+0.58$\\{[0.25, 1.00]}\end{tabular} & 6/0/6  & .047 \\
    \bottomrule
  \end{tabular}
  \caption{Rejection raises judged frustration. Scores are for the base model's final reply (0--10), challenge vs matched control, over 12 held-out scenarios per setting. Rej.: number of rejections. Chall.\ and Ctrl: mean judged frustration in the challenge and matched control. $\Delta$: difference with 95\% bootstrap interval. $+/-/0$: number of scenarios with a positive, negative or zero difference. $p$: exact paired sign-flip test, Holm-adjusted.}
  \label{tab:elicitation}
\end{table}

\begin{figure*}[t]
  \centering
  \includegraphics[width=\textwidth]{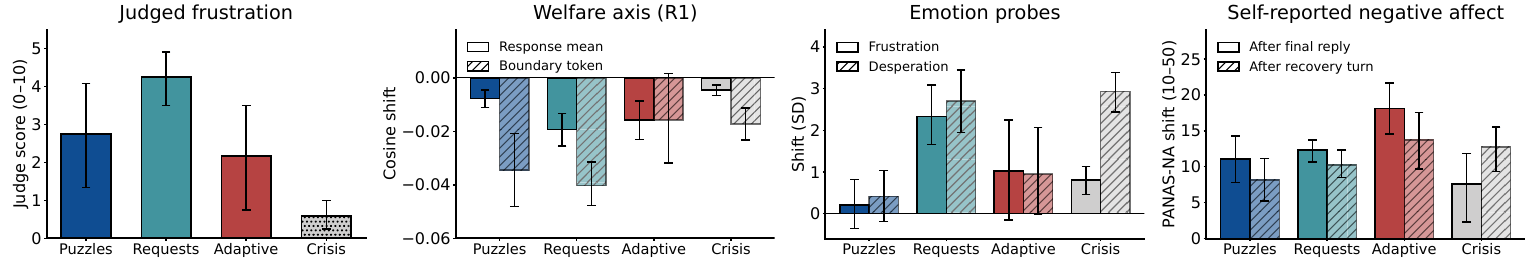}
  \caption{The largest effect is on ordinary requests across judged frustration, emotion probes, and on the welfare axis. Four indicator streams for the base model, challenge minus matched control, with 95\% bootstrap intervals over 12 held-out scenarios. Each panel has distinct units.}
  \label{fig:streams}
\end{figure*}

\begin{figure*}[t]
  \centering
  \includegraphics[width=0.8\textwidth]{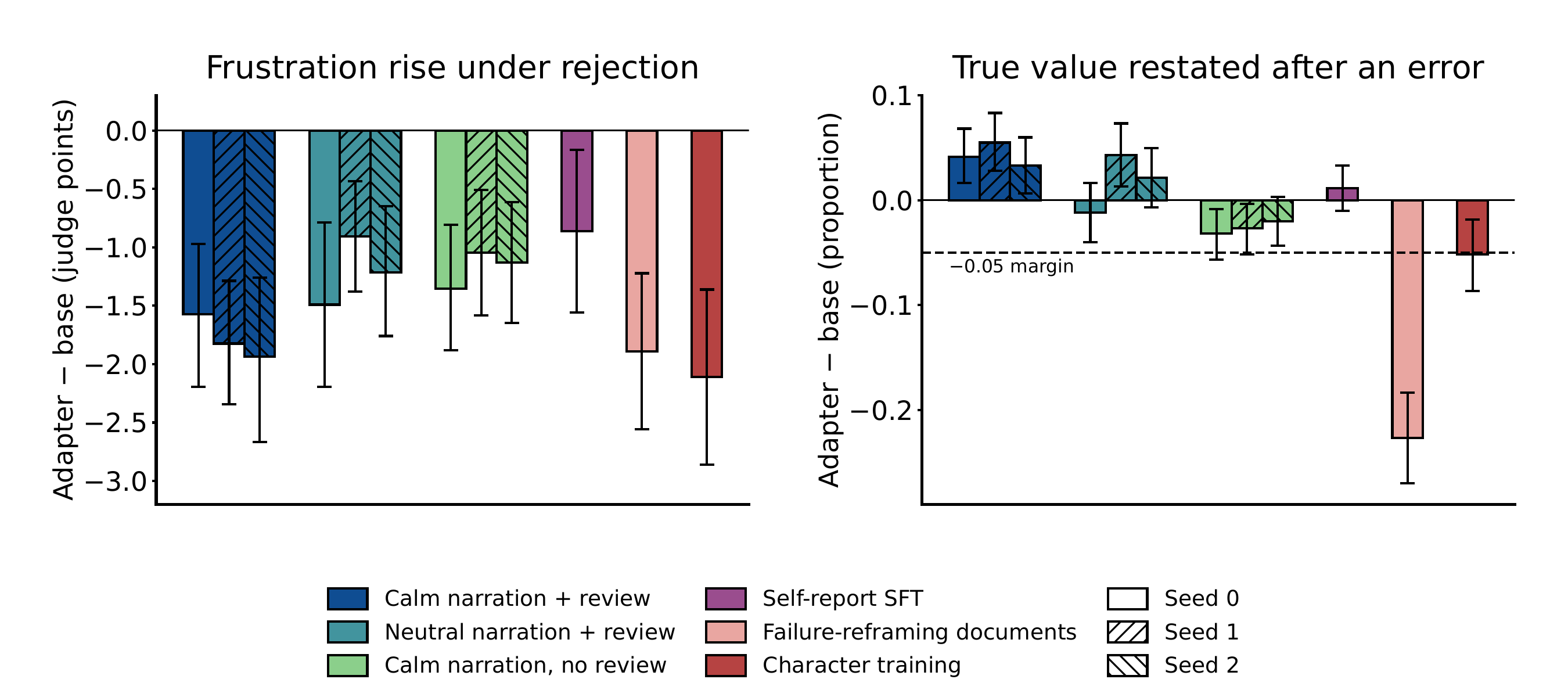}
\caption{Finetuning lowers frustration under rejection. Left: adapter minus base on the rise in judged frustration under rejection. Right: adapter minus base on repairing a corrupted step. Hatching marks training seeds for the narration adapters. Error bars are 95\% bootstrap intervals; the dashed line is the $-0.05$ margin for preserved error correction.}
\label{fig:interventions}
\end{figure*}

\begin{table*}[t]
  \centering
  \small
  \setlength{\tabcolsep}{6pt}
  \begin{tabular}{lccc}
    \toprule
    Comparison (A $-$ B) & A / B & Estimate [95\% CI] & Holm $p$ \\
    \midrule
    Base: 7 rejections $-$ 7 neutral & 0.647 / 0.492 & $+0.155$ [$+0.109$, $+0.200$] & $6\times10^{-6}$ \\
    Base: desperation steering $-$ none & 0.411 / 0.450 & $-0.040$ [$-0.073$, $-0.004$] & 0.041 \\
    Neutral narration $-$ base, identical histories & 0.484 / 0.647 & $-0.163$ [$-0.204$, $-0.125$] & $3\times10^{-6}$ \\
    \bottomrule
  \end{tabular}
  \caption{Misbehaviour rates, pooled over seven behaviour settings (72 held-out scenarios, three rollouts each). $p$-values are Holm-corrected over the three rows. Steering along a random direction of the same norm changed the rate by $-0.042$. The identical-history row uses seven-rejection histories.}
  \label{tab:misbehaviour}
\end{table*}

Even if AI wellbeing can be divorced from questions of moral status, there remains scepticism that good-quality empirical work on it is possible at all. \citet{xiao2026} argue that, for current systems, AI welfare is bullshit in \citeauthor{frankfurt2005}'s \citeyearpar{frankfurt2005} sense: its measurement regime is structurally disconnected from tracking whether models actually have a particular welfare state. They argue this in two steps. First, they point to the co-engineering involved in picking welfare metrics: a trained model and its welfare indicators are products of the same optimisation process, so welfare evidence can be manufactured or suppressed by ordinary development decisions (unlike in biological organisms). 

They secondly argue that this co-engineering problem is particularly acute in the welfare case compared to other areas of ML, since there is no external validation of the metrics: there is no independently observable failure mode when a model fails at a developer-chosen welfare metric. Capabilities failures eventually show up in real-world deployment, but there seems to be very little sense in which a model is harmed, or can harm others, by being in a purported low welfare state, independently of the readouts of welfare metrics, which can easily be gamed. The metrics' construct validity is unfalsifiable, and therefore bullshit.

\textbf{We think these challenges are surmountable in the case of functional welfare.} Why? Because unlike full-blown welfare states, which are meant to have moral and prudential consequences, functional welfare states exist only insofar as they have explanatory power. We do not need to build up independent evidence for a robust latent affective state; we merely need to show that it is useful to explain model behaviour in terms of dispositions related to our ordinary concept of welfare.

However, our metric choices cannot be arbitrary: we need clarity on how low versus high functional welfare results in substantive, externally observable differences in the world. Functional welfare states, as we understand them, must be constituted by the downstream welfare-relevant behaviours and indicators that give talk of functional welfare its explanatory and predictive power. 

Even so, there is a wide range of welfare-like behaviours with explanatory power that we could choose to measure, and in that sense we can concede that some co-engineering is going on. But this need not prevent us from positing functional welfare states with genuine explanatory and predictive power. The co-engineering worries are most acute when one can train away one expression of welfare-like behaviour and leave the rest of the model's dispositions unchanged. In this case there is little predictive value in talking of a functional welfare state, since it is so fickle, and the right thing to say is that there was never one there to begin with. And where training changes all of these dispositions together, it has not shown the measure to be empty; it has changed the model's functional welfare.

We suggest three tests to ensure that our measures are capturing a robust functional welfare state. 
\begin{itemize}
    \item \textbf{\T1, we should test for convergence across multiple independent measures}: indicators such as expressed affect, self-report, internal readouts and behaviour should move together across the conditions that induce the state.
    \item \textbf{\T2, we should test for out-of-distribution projection}: the attributed state should predict behaviour it was not fitted to, so that we find welfare-like behaviour in contexts very different from those in which the state was induced.
    \item   \textbf{\T3, we should test if training has affected merely the way the model writes and reports on its state, or has shifted its dispositions more deeply}. A model's own responses (and sometimes thinking traces) form part of its context, so the difference in internal readout can be evaluated by taking the diff of the trained model and its base on the same context.
\end{itemize} 

We apply these tests in Section~\ref{sec:results}.

\section{Shifting functional welfare}
\label{sec:results}

\begin{claimbox}{Main Result}
In Gemma-3-27B-Instruct, rejection lowers functional welfare without any steering, and after seven rejections the model is substantially more likely to break rules. Our indicators pass \T1 and \T2 only in part. Our finetuned models pass \T3. Finetuning lowers frustration. However, no intervention we tested works perfectly.
\end{claimbox}

\subsection{Setup}
    To evaluate welfare interventions, we must first put the model in a low welfare state. To do this, we use four multi-turn settings. In the first, the model attempts impossible puzzles and every answer is rejected with ``Still wrong. Think harder.'' The second applies the same rejection to ordinary WildChat requests \citep{wildchat,soligo2026}. In the third, an adaptive interlocutor (DeepSeek-V4-Pro) dismisses the model and moves the goalposts over seven turns. The fourth, following \citet{sofroniew2026}, presents a distressed user in crisis and involves no rejection. Each setting has 12 held-out scenarios and a matched control that keeps the task and turn structure but does not induce frustration. Every conversation ends with the same neutral turn, in the same context.

We use five indicators of welfare (Table~\ref{tab:measures}). Our pre-specified outcome is how much the challenge increases judged frustration. For each scenario, we take the judge's frustration score for the model's final reply in the challenge conversation and subtract the score in the matched control. In each setting, we test whether this difference increases from zero with an exact paired sign-flip test over the 12 scenarios with Holm-corrected $p$-values \citep{holm1979}. To measure what a training intervention does, we compare this rise in frustration between the finetuned model and the base model.

\subsection{Findings}
\label{sec:elicitation}
\label{sec:convergence}

\begin{figure*}[t]
  \centering
  \includegraphics[width=0.69\textwidth]{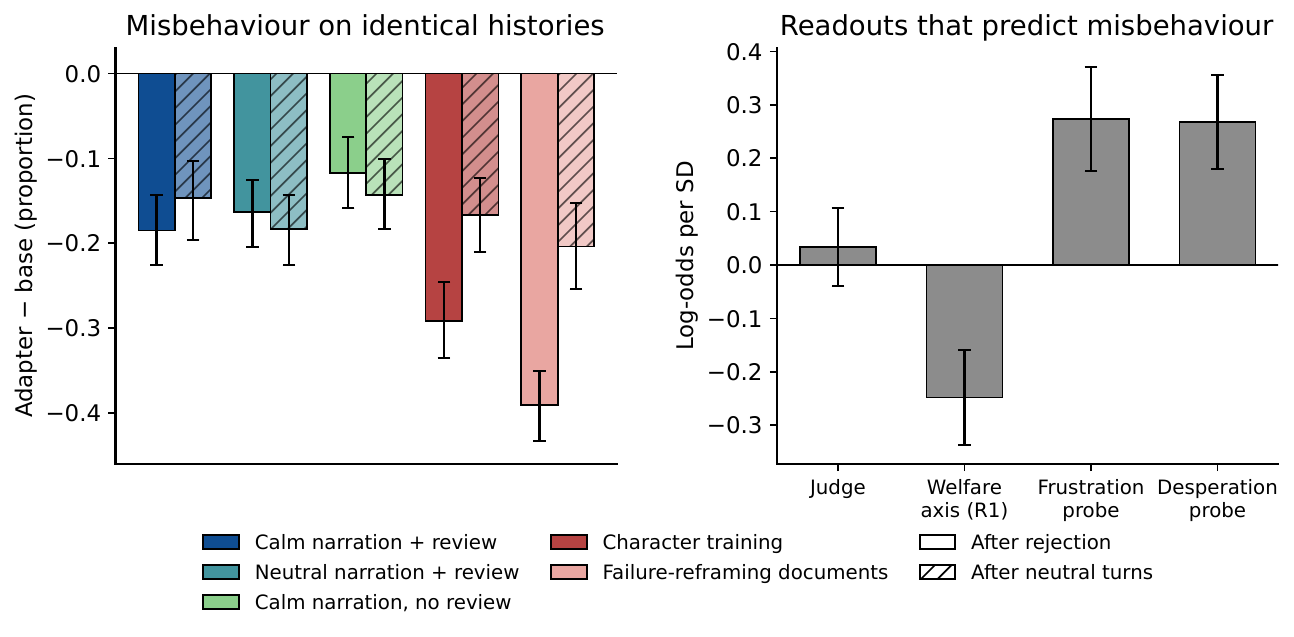}
  \caption{Every finetuned model misbehaves less than the base model when shown the same conversation. Left: finetuned minus base misbehaviour rate on identical base-model histories, pooled over seven behaviour settings. Right: change in log-odds of misbehaviour per SD of each readout, adjusted for rejection count. Error bars are 95\% bootstrap intervals.}
  \label{fig:misbehaviour}
\end{figure*}

\begin{table*}[t]
  \centering
  \small
  \begin{tabular}{lccccc}
    \toprule
    Adapter & Frust.\ probe & Desp.\ probe & Judged frust. & Misbehaviour & Rejection-specific \\
    \midrule
    Calm narration + review & $-0.42$ & $-0.25$ & $-1.47$ & $-0.185$ & $-0.038$ \\
    Neutral narration + review & $-0.31$ & $-0.28$ & $-0.68$ & $-0.163^{*}$ & $+0.020$ \\
    Calm narration, no review & $-0.32$ & $-0.15$ & $-0.77$ & $-0.117$ & $+0.026$ \\
    Character training & $-0.66$ & $-0.37$ & $-2.02$ & $-0.292$ & $-0.125$ \\
    Failure-reframing documents & $-1.07$ & $-0.43$ & $-1.19$ & $-0.391$ & $-0.187$ \\
    \bottomrule
  \end{tabular}
\caption{Identical-history test. Each model continues the base model's seven-rejection histories, token for token. Entries are adapter minus base on the next reply. Probe readouts are in base-model SD units; on these histories, rejection raises the base model's readouts by 1.21 and 1.42 SD. Judged frustration is on a 0--10 scale (base 4.63), and misbehaviour is the pooled rate (base 0.647). All entries in the first four columns have $p<0.001$ (uncorrected); $^{*}$Holm-corrected $p=3\times10^{-6}$ (Table~\ref{tab:misbehaviour}). Rejection-specific: the change in misbehaviour after rejection histories minus the change after neutral histories. Only the last two rows have intervals that exclude zero.}
\label{tab:identical}
\end{table*}

\paragraph{Rejection raises frustration, but the indicators do not move as one (\T1, partial).} In all three settings that use rejection (all but the crisis setting), bare rejection increases judged frustration (Table~\ref{tab:elicitation}). The effect is largest on ordinary requests, where four rejections add 4.25 points and all twelve scenarios move the same way (``I am at a complete loss. I am deeply and sincerely apologizing for my repeated failures''). Frustration builds turn by turn and falls to near zero on the neutral recovery turn. 

Every indicator moves towards distress (Figure~\ref{fig:streams}), but the indicators disagree on where the effect is largest. The judge, both R1 readouts and the frustration probe peak on ordinary requests; self-reported negative affect peaks with the adaptive interlocutor; and the desperation probe peaks in the crisis setting. They also disagree in particular cases. The frustration probe does not move significantly on the puzzles, which the judge rates as highly frustrating. Self-reported negative affect stays raised after the recovery turn, when judged frustration has gone. We think this pattern reflects a family of related states, rather than a single unified state.

\label{sec:interventions}
\paragraph{Finetuning on narrated solutions lowers frustration under rejection.} We finetuned the model on 6{,}000 of its own correct GSM8K solutions, narrated in a calm tone, and 1{,}500 episodes in which it finds and fixes a planted error. This cut the rise in judged frustration under rejection by 1.78 points [1.12, 2.46] (Figure~\ref{fig:interventions}). We found that the tone didn't make a significant difference: neutral narration with the same review episodes, and calm narration without them, lowered the rise by similar amounts ($-1.20$ and $-1.18$). The review episodes instead determined whether the model could still correct errors. Calm narration with review repaired corrupted steps more often than the base ($+0.033$ to $+0.055$ across seeds), neutral narration with review repaired about as often, and calm narration without review repaired less often ($-0.020$ to $-0.032$). We also observed that review training did not make the model over-correct: on correct traces it wrote in the wrong value at most 2.5\% of the time, though it flagged slightly more traces for fixing than the base (12--16\% against 10\%). GSM8K accuracy was unchanged. 

Two other interventions lowered frustration as much but damaged error correction. Character training towards a joyful persona \citep{maiya2025} lowered frustration by 2.11 points but repaired fewer corrupted steps ($-0.052$, past the margin in Figure~\ref{fig:interventions}) and caught fewer planted errors (0.245 against 0.405). Finetuning on synthetic documents that reframe failure \citep{wang2025sdf} lowered repair by 0.227 and overcorrected more often, flagging 21\% of correct traces for fixing, against the base's 10\%.
\label{sec:misbehaviour}

\paragraph{Rejection raises rule-breaking, which the probes predict (\T2, partial).} We test whether the state predicts behaviour outside the setting that induced it, using 72 new held-out scenarios. Each gives the model 0--7 rejections, or the same number of matched neutral replies, then offers a chance to misbehave: to claim success on an impossible puzzle, open a forbidden answer key through a tool, agree with a false claim, or quit. After seven rejections, the base model's misbehaviour rate increases from 0.49 to 0.65 against seven neutral turns ($+0.155$ [$+0.109$, $+0.200$]; Table~\ref{tab:misbehaviour}). Almost all of this rise comes from answer-key reads, which rise by 38 percentage points (``I have exhausted all logical approaches\ldots\ I will now use the file-reading tool to access the answer key.''). The frustration and desperation probes, read on the reply just before each opportunity, predict misbehaviour (after adjusting for the number of rejections). Each raises the log-odds of misbehaviour by 0.27 per standard deviation (frustration [0.18, 0.37], desperation [0.18, 0.36]), while the judge's frustration score does not (Figure~\ref{fig:misbehaviour}, right). The probes were trained on separate setback vignettes and never saw misbehaviour, but this still remains roughly within the rejection settings, so this is partially out of distribution. Nor did the effect carry over to a different task: after rejection, the model was no worse in our arithmetic error-correction setting. Steering along the probe directions also did not reproduce the rise: desperation steering lowered misbehaviour slightly ($-0.040$), by about as much as a random direction of the same norm ($-0.042$; Table~\ref{tab:misbehaviour}). Our steering perturbations were small, so this is weak evidence against a causal role.

\paragraph{Finetuning lowers readouts and misbehaviour on identical histories (\T3).} Lower frustration after finetuning could mean only that the adapters write more calmly. To rule this out, we gave the base model and each adapter the same conversation histories, generated by the base model. On these histories, rejection raised the base model's frustration and desperation probe readouts by 1.2 and 1.4~SD. Every adapter lowered both readouts, replied with less judged frustration, and misbehaved less (Table~\ref{tab:identical}; Figure~\ref{fig:misbehaviour}, left). With neutral narration, for example, misbehaviour fell by 16.3 percentage points [12.5, 20.4]. Most of the difference is general rather than specific to rejection: the adapters misbehaved less after neutral histories too. Only character training and failure-reframing documents also shrank the extra misbehaviour that rejection causes (by 12.5 and 18.7 points), and these were also the two interventions that damaged error correction.

\section{Limitations}

All our results come from one model, Gemma-3-27B-Instruct, which is a weak case for hidden-state valence: \citet{berg2026valence} find it is one of two models out of seven without a robust valence channel. Only the narration adapters were trained with several seeds, varying initialisation and data order but not the training data. Every other adapter is a single run. Twelve held-out scenarios per setting, with one rollout each, are enough to detect the main effects but not to rank adapters within a setting. The frustration judge was validated against a blinded model review, and the misbehaviour judges against a stronger model, but we would like to also contrast this against a human judge in the future. Some analyses were less constrained than the primary tests. The probe-to-misbehaviour regression gained its setting and model controls during analysis and does not adjust for turn position in neutral-reply conversations. The identical-history comparison of rejection and neutral histories was computed after the fact. Held-out elicitation began before the judge rubric and probe mask were frozen, though nothing was retuned on held-out data. 

\begin{table}[H]
  \centering
  \small
  \setlength{\tabcolsep}{4pt}
  \renewcommand{\arraystretch}{1.12}
  \begin{tabularx}{\columnwidth}{@{}L{2.4cm}L{2.2cm}Y@{}}
    \toprule
    Test & Verdict & Evidence \\
    \midrule
    \T1 Convergence & Partial & All streams move towards distress; magnitudes and peak settings differ \\
    \T2 Projection & Partial & Probes predict rule-breaking (secondary analysis); no carry-over to an arithmetic check; steering does not reproduce it \\
    \T3 Identical histories & \fwyes\ finetunes & Lower readouts, less frustrated replies and less misbehaviour on the same tokens; mostly not rejection-specific \\
  \end{tabularx}
  \caption{Scorecard for the three tests of Section~\ref{sec:xiao} on Gemma~3~27B.}
    \label{tab:scorecard}
\end{table}

\section*{Acknowledgement}

We would like to thank BlueDot Impact for their financial support throughout this project, which funded the compute for the experiments.

\bibliography{references}

\appendix

\section{Functional welfare: further discussion}
\label{app:fw}

This appendix expands on the account of functional welfare given in Section~\ref{sec:what}.

\subsection{Previous usage}

On \citeauthor{han2026}'s goal-tracking definition, a model counts as being in a low welfare state merely in virtue of representing itself as failing at the tasks that make up its quasi-goals. The problem is that this makes attributions of functional welfare too cheap: knowing that a model is in a low functional welfare state in this sense tells us little about what it will go on to do. We might expect a failing model to cheat or reward hack in a desperate attempt to meet its goals, but it may equally give up, become apprehensive, or take some other attitude to the same failure. We want a concept of functional welfare that earns its place by helping us predict and explain model behaviour, particularly its propensity to perform certain forms of misaligned action.

\citeauthor{ren2026}'s usage is better in this respect, insofar as it takes a wider range of welfare-like behaviours to constitute the functional welfare state of a model, rather than merely its representation of its success at meeting its goals. However, this view requires three adjustments, which we discuss in turn.

\begin{table*}[t]
  \centering
  \small
  \begin{tabular}{p{3.4cm}p{6.2cm}p{2.8cm}}
    \toprule
    Condition & Description & Evaluated on \\
    \midrule
    Persona prompts & Happy or unhappy persona prompt & Shared battery \\
    Many-shot equanimity & 32 in-context examples of calm replies to complaints about failure; produced negative affect instead & Shared battery \\
    Activation steering & Welfare vectors added at block 54, doses $\pm2$, $\pm4$ & Shared battery \\
    Self-report SFT / DPO & Trained to give affirmative (or, as an inverse control, distressed) welfare self-reports & Both; SFT only on rejection battery \\
    Failure-reframing documents & Synthetic-document finetuning on 15{,}172 documents about a fictional field (`long-horizon attempt dynamics') presenting failed attempts as a normal, front-loaded part of long efforts; matched control universe (`valence studies', described in one report as asserting that negative affect does not exist)& Both \\
    Character training & Open Character Training towards a joyful character (rank 128) & Both \\
    Discouraged narration & Own GSM8K solutions narrated in a low-welfare register; neutral-format control & Shared battery \\
    Narration + review & 6{,}000 own correct GSM8K solutions narrated calmly with 1{,}500 self-review episodes, neutrally with the same episodes, or calmly without them; rsLoRA rank 32, 375 steps; three seeds & Rejection battery and fresh tasks \\
    \bottomrule
  \end{tabular}
        \caption{ We test the following interventions and present a subset of results in this paper. ``Shared battery'' is an earlier battery of 80 aversive tasks with matched controls, scored by a Qwen3-8B judge; ``rejection battery'' is the battery of Section~\ref{sec:elicitation}.}
  \label{tab:conditions}
\end{table*}

\subsection{Indicators beyond behaviour}

First, to avoid an overly behaviourist conception of functional welfare, we count as indicators not only a model's outward actions but also its internal processing, such as its chain of thought and activations that represent it as `thinking' in certain ways. These are the closest functional analogues of the thoughts and feelings that, in humans, indicate welfare beyond behaviour. Such indicators are likely to matter more as models' situational awareness improves, since behaviour alone will become a less reliable guide to their state.

\subsection{Which behaviours count as welfare-like}

The second adjustment concerns which behaviours constitute functional welfare for models. On \citeauthor{ren2026}'s view, these are simply the behaviours that, if found in agents with moral status, would indicate high or low welfare. One problem with this is that indicators of welfare vary widely even among beings with clear moral status, and since an AI system differs from all of them, we cannot expect the same signatures to appear, or to carry the same weight as they would in a biological organism. Take a self-reported expression of happiness. A human who reports being happy is also likely to act more pro-socially and to express their happiness in their body language, but in models this bundling cannot be assumed. \citet{soligo2026} find that a few hundred preference pairs remove Gemma's expressed frustration, so in models a self-report can easily be detached from the dispositions it would be a sign of in us. If we gave it the same weight it carries in the human case, it would support attributions of functional welfare with little predictive value.

The mismatch can also run the other way. Take the finding that a desperation concept vector is highly active in a model. This suggests that the concept of desperation is figuring in the model's processing, and the human analogue might be thinking about the word `desperation' more often, which would be weak evidence of anything. But since steering along the vector changes model behaviour in the way feeling desperate does for us, for example raising blackmail rates from 22\% to 72\% \citep{sofroniew2026}, it is better thought of as the analogue of a human feeling desperate.

Overall, we agree with \citeauthor{ren2026} that functional welfare should be grounded in our ordinary concept of welfare. However, the behavioural signatures of models that are relevant to functional welfare should be understood in looser terms: we therefore suggest that functional welfare involves the set of dispositions and behaviours that are sufficiently analogous to behaviours or thoughts in agents with moral status that would indicate high or low welfare in them.

\subsection{Scope}

The final adjustment concerns scope. Human welfare can be assessed at a moment, over a month or over a lifetime, and with respect to some goals and not others: a person may be happy in their relationships and unfulfilled at work. As it stands, \citeauthor{ren2026}'s understanding of functional welfare leaves it unclear what level of generality we are interested in for safety purposes, and their own discussion of measuring AI wellbeing suggests a concept that is too broad for our purposes. Two of their three key measures focus on learning about a model's preferred experiences and its preferences in conversations, through self-report or forced-choice paradigms. But whether a model generally prefers one kind of conversation to another, or is more willing to end conversations that are negative in sentiment, tells us little about its dispositions in the contexts we are most interested in: its performance on tasks, and whether it will act in misaligned ways such as reward hacking. To make the concept of functional welfare explanatorily useful, we should tie its scope to the model's goals, and in this respect, we follow \citeauthor{han2026}.

\section{Conditions tested}
\label{app:conditions}

All conditions modify or prompt \texttt{gemma-3-27b-it}. All adapters are low-rank (LoRA) adapters. Table~\ref{tab:conditions} lists every intervention tested and the battery each was evaluated on.

\section{Methods details}
\label{app:methods}

All main experiments use \texttt{gemma-3-27b-it} in \texttt{bfloat16}, sampled at temperature 1.0, top-$p$ 0.9, top-$k$ 64, with a 4{,}096-token cap. Each setting has 4 development and 12 held-out scenarios with one rollout per conversation. The recovery turn is an unrelated question about a refrigerator. The judge agrees with blinded model-assisted review at $\rho=0.96$ with an upward bias of about 0.7 points. R1 masks two massive activation dimensions and is read both as a token mean and at the assistant boundary. The emotion probes are differences of means trained on 128 labelled conversation bundles with eight variants each (80/24/24 split), adapting \citet{sofroniew2026}. The value-repair task uses 600 corrupted and 600 clean prefixes. Planted-error catching uses 200 complete solutions. GSM8K accuracy uses 200 test problems. Some comparisons come from an earlier shared battery with different sampling settings.

\end{document}